\documentclass[10pt,twocolumn,letterpaper]{article}

\usepackage{iccv}
\usepackage{times}
\usepackage{epsfig}
\usepackage{graphicx}
\usepackage{adjustbox}
\usepackage{amsmath}
\usepackage{amssymb}
\usepackage{gensymb}
\usepackage{algorithm}
\usepackage{algpseudocode}

\usepackage[pagebackref=true,breaklinks=true,letterpaper=true,colorlinks,bookmarks=false]{hyperref}
\usepackage{url}

\iccvfinalcopy

\def\iccvPaperID{18}

\ificcvfinal\fi

\begin{document}

\title{BuilDiff: 3D Building Shape Generation using \\Single-Image Conditional Point Cloud Diffusion Models}

\author{Yao Wei\\
University of Twente\\
The Netherlands\\
{\tt\small yao.wei@utwente.nl}
\and
George Vosselman\\
University of Twente\\
The Netherlands\\
{\tt\small george.vosselman@utwente.nl}
\and
Michael Ying Yang\\
University of Twente\\
The Netherlands\\
{\tt\small michael.yang@utwente.nl}
}

\maketitle

\begin{figure*}[t!]
    \centering
    \includegraphics[width=0.77\linewidth]{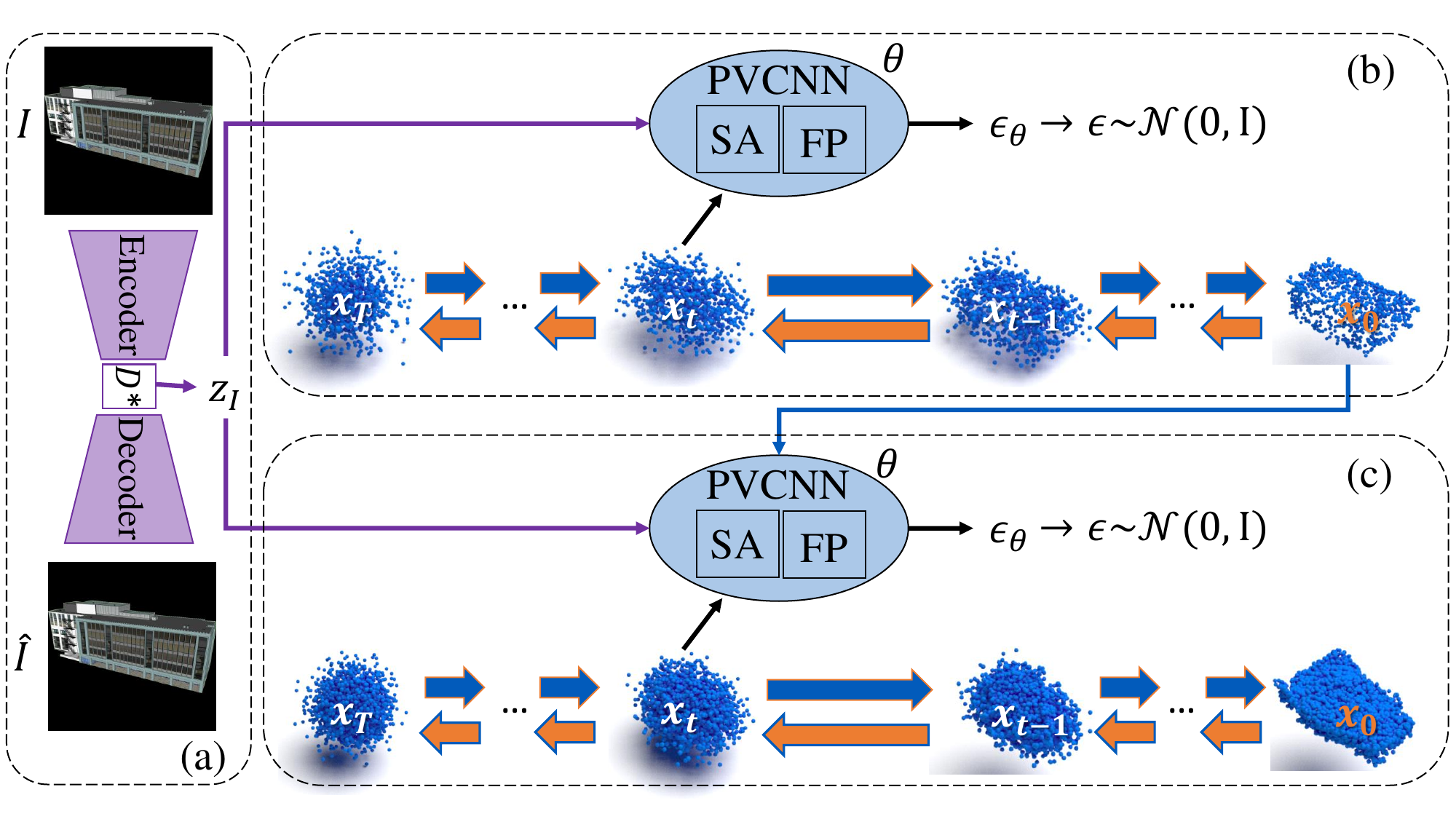}
    \caption{Overall pipeline of BuilDiff. (a) First, a CNN-based image auto-encoder is pre-trained to derive image embedding $z_I$ in the latent space. (b) Second, a point cloud diffusion model $\theta$ conditioned on $z_I$ is trained for denoising from noise, i.e., $x_T$, to a desired shape $x_0$. (c) Finally, another point cloud diffusion model is trained to sample a high-resolution shape conditioned on $z_I$ and the low-resolution shape inferred by the previous diffusion model.}
    \vspace{-3mm}
    \label{fig:pipeline}
\end{figure*}

\begin{abstract}
3D building generation with low data acquisition costs, such as single image-to-3D, becomes increasingly important. However, most of the existing single image-to-3D building creation works are restricted to those images with specific viewing angles, hence they are difficult to scale to general-view images that commonly appear in practical cases. To fill this gap, we propose a novel 3D building shape generation method exploiting point cloud diffusion models with image conditioning schemes, which demonstrates flexibility to the input images. By cooperating two conditional diffusion models and introducing a regularization strategy during denoising process, our method is able to synthesize building roofs while maintaining the overall structures. We validate our framework on two newly built datasets and extensive experiments show that our method outperforms previous works in terms of building generation quality. 

\end{abstract}

\vspace{-3mm}
\section{Introduction}

Buildings play a role in various applications including urban modeling \cite{shiode20003d}, industrial design \cite{reyes20173d}, and virtual reality \cite{wang2010new}. Since buildings usually have far more complicated 3D shapes than the clean models in common 3D object datasets, e.g., ShapeNet \cite{chang2015shapenet}, efficient acquisition of 3D buildings remains an open problem in both synthetic and real scenarios. On one hand, designers typically use CAD technologies to create crafted 3D building models for synthetic scenarios (e.g., games and films), which inevitably requires tremendous manual effort. On the other hand, 3D shapes of real-world buildings are mostly acquired using mobile platforms such as airplanes, which are equipped with LiDAR sensors that capture the 3D coordinates of observed objects. Despite the recent release of several 3D city models where buildings are the primary components \cite{hu2021towards,peters2022automated}, obtaining high-quality 3D building models is still a costly process, leading to limited availability in many parts of the world. Therefore, there is an urgent need for automated methods to generate realistic 3D buildings.

As an alternative to LiDAR point clouds, optical images obtained by regular cameras that have a lower acquisition cost, capture rich information about buildings as well. Over the past few decades, image-based 3D building reconstruction has long  been a research focus \cite{torii2009google,duan2016towards,yu2021automatic}. Most of the existing works rely on photogrammetric technologies (e.g., Structure-from-Motion) leveraging multiple overlapping images taken from different viewpoints. Such requirements, however, restrict the applicability of those approaches to certain situations where multi-view images are infeasible to obtain. 
In contrast, single-image 3D reconstruction \cite{mahmud2020boundary,zhao2021shape, pang20223d} offers a low-cost solution towards efficient 3D building shape generation, which could reduce the time-consuming manual design of synthetic buildings and facilitate the digital simulations of real buildings. 

Driven by deep learning techniques, generative models \cite{kingma2013auto,goodfellow2014generative,rezende2015variational,ho2020denoising} have shown promising results on 3D computer vision, e.g., single image-to-3D point cloud \cite{fan2017point,gadelha2018multiresolution,wei2022flow}. However, these methods are mainly designed and trained on synthetic and symmetric 3D objects like airplanes, and they have limited applicability when generating 3D building models with complex structures in real-world scenarios. Besides, it is observed that single image-to-3D building methods \cite{srivastava2017joint,mahmud2020boundary,li20213d} mostly require remote sensing images taken from specific perspectives, e.g., nadir-view, and the generated building models exhibit low level-of-details (LoDs) \cite{peters2022automated} without roof structures.

In this work, we present \textbf{BuilDiff} that takes a step into single general-view image-to-3D \textbf{buil}ding synthesis by leveraging \textbf{diff}usion models in a coarse-to-fine manner. 

Our \textbf{main contributions} are summarized as follows:
\begin{itemize}
    \item A novel hierarchical framework named \textit{BuilDiff} is proposed to generate realistic 3D shapes of buildings with roof structures, i.e., at LoD2, given their single general-view images. 
    \item Guided by an image auto-encoder, a base diffusion model coarsely identifies the overall structures of buildings, and a upsampler diffusion then derives higher resolution point clouds.
    \item A weighted building footprint-based regularization loss is introduced to constrain building structures and avoid ambiguous guidance during denoising process.

\end{itemize}
   
Experiments demonstrate the effectiveness of the proposed method on synthetic and real-world scenarios.

\section{Related Work}

\noindent
\textbf{Single Image 3D Reconstruction.}
Reconstructing the 3D shape of an object from its single-view image is a long-standing research problem in computer vision \cite{choy20163d,fan2017point}. This is motivated by human perception: given a single object image, humans can infer its potential 3D structure based on the cue image as well as prior knowledge of the 3D world. Many deep learning-based methods \cite{gadelha2018multiresolution,xu2019disn,wen20223d,wei2022flow} explore such prior knowledge by learning from abundant pairs of 3D shapes and single images. However, most of these methods concentrate on 3D benchmarks such as ShapeNet \cite{chang2015shapenet}, which are restricted to synthetic objects with simple and symmetric 3D shapes. To capture the complexity of real-world objects, recent works \cite{duggal2022topologically,melas2023pc} exploit camera poses of images to reconstruct 3D shapes that are well aligned with the input images. While the camera poses contribute to 2D-3D correspondences, this formulation usually suffers from inflexibility for images without pose information. There can potentially be multiple solutions for pose estimation, but an inaccurate pose may result in low-resolution geometries. 

Unlike common objects, buildings have much more challenging structures \cite{selvaraju2021buildingnet}. For real-world 3D building reconstruction from a single remote sensing image, existing works \cite{srivastava2017joint,mahmud2020boundary,li20213d} roughly regard it as a combination of building footprint extraction and height prediction from nadir-view images. To generate more detailed 3D buildings, \cite{pang20223d} performs image-to-mesh by leveraging street-view images aided by nadir-view images, and \cite{zhao2021shape} develops implicit representations of object-level buildings. A recent work \cite{rezaei2022sat2pc} specifically focuses on generating 3D point clouds of building roofs. Although these methods could generate relatively simple buildings, they fail to deal with complex buildings. Moreover, the applicability remains limited to view-specific input images, and general-purpose single-image 3D building reconstruction is less explored.

\begin{figure*}[t!]
    \centering
    \includegraphics[width=0.86\linewidth]{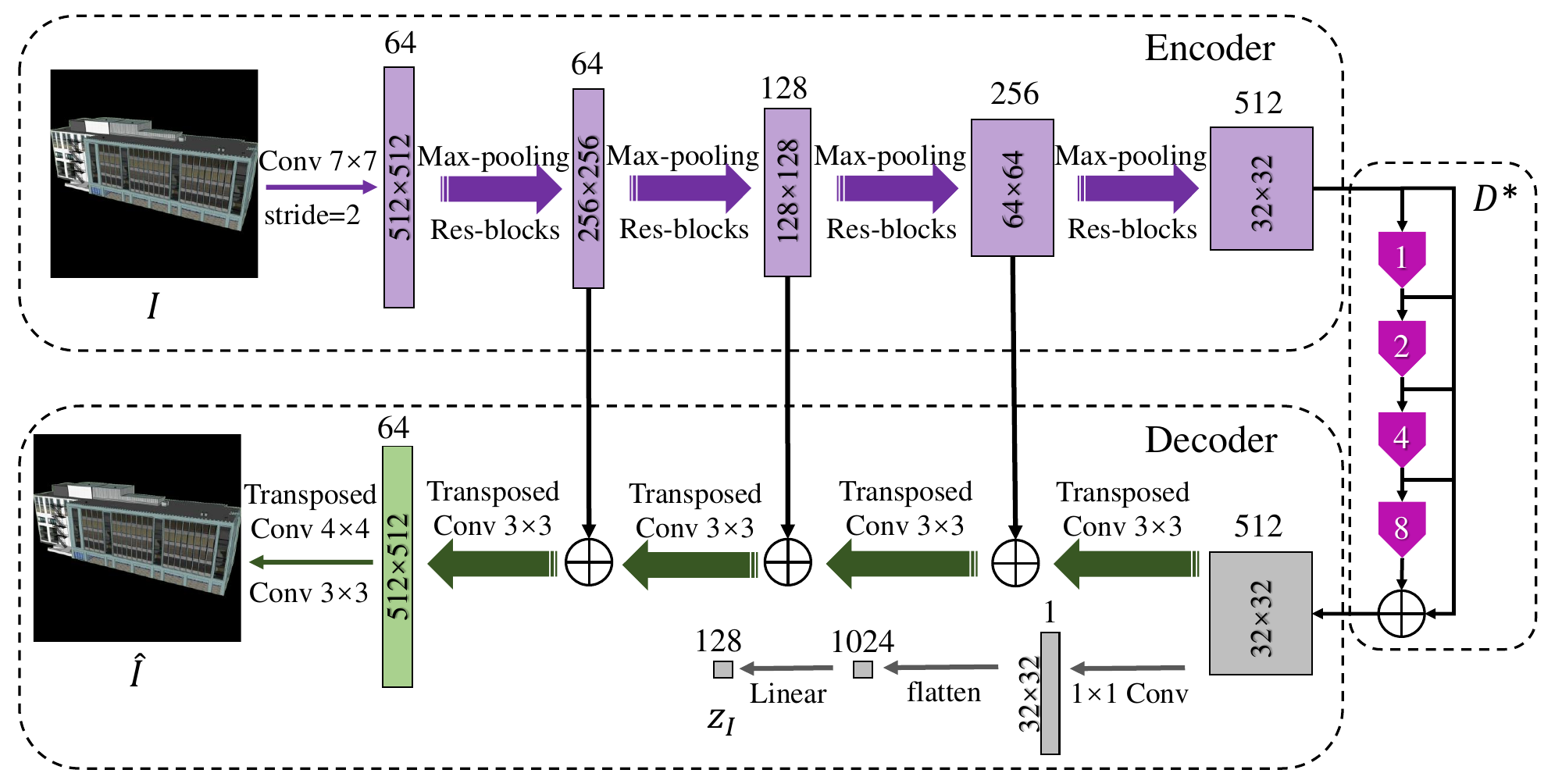}
    \caption{The architecture of image auto-encoder network. The input is RGB image $I\in\mathbb{R}^{H\times W\times3}$, and the output contains reconstruction $\hat{I}\in\mathbb{R}^{H\times W\times3}$ and an image embedding $z_I\in\mathbb{R}^{d}$. The channel numbers and sizes of the feature maps are annotated above and inside them, respectively. $\oplus$ indicates addition process, and the black arrows between encoder and decoder are skip connections.}
    \vspace{-4mm}
    \label{fig:supp_ae}
\end{figure*}

\vspace{1mm}
\noindent
\textbf{Deep Generative Models for 3D Point Cloud.}

In recent years, many generative models have been proposed and developed, including Variational Autoencoders (VAE) \cite{kingma2013auto}, Generative Adversarial Networks (GANs) \cite{goodfellow2014generative}, Normalizing Flows (NFs) \cite{rezende2015variational}, and Diffusion models \cite{ho2020denoising,rombach2022high}, which have achieved impressive results on data synthesis, especially for image synthesis tasks \cite{ramesh2022hierarchical}. With great success in the 2D domain, deep generative models are gradually applied in the 3D domain. Besides using GANs \cite{achlioptas2018learning} and NFs \cite{yang2019pointflow}, a growing number of works \cite{luo2021diffusion,zhou20213d,zeng2022lion,nichol2022point,melas2023pc} leverage diffusion models for 3D point cloud synthesis. 

Parameterized by two Markov chains, denoising diffusion probabilistic models (DDPMs) \cite{ho2020denoising} consist of a forward process and a reverse process. The former gradually adds random noise to input data $x_0$, and the latter reconstructs $x_0$ starting from the random noise $x_T$ where $T$ indicates the total number of time steps. Compared with other generative models, denoising diffusion models demonstrate the advantages of high quality and diversity, which offer a promising avenue for the 3D shape synthesis of complex objects such as buildings. Hence, we perform 3D building reconstruction from a single image by scaling diffusion models from unconditional to conditional settings. To decrease the diversity while increasing the quality of each individual sample, two types of guidance are presented for diffusion models: classifier guidance \cite{dhariwal2021diffusion} and classifier-free guidance \cite{ho2022classifier}. 

\section{Method}

Instead of only exploiting images from specific perspectives, e.g., nadir- or street-views, our goal is to generate 3D point clouds of buildings from a single general-view image, aiming to improve the applicability of the proposed method. As shown in Fig. \ref{fig:pipeline}, we introduce a hierarchical framework BuilDiff which consists of three components: (a) image auto-encoder, (b) image-conditional point cloud base diffusion, and (c) image-conditional point cloud upsampler diffusion. 

\begin{figure*}[t!]
    \centering
    \includegraphics[width=.88\linewidth]{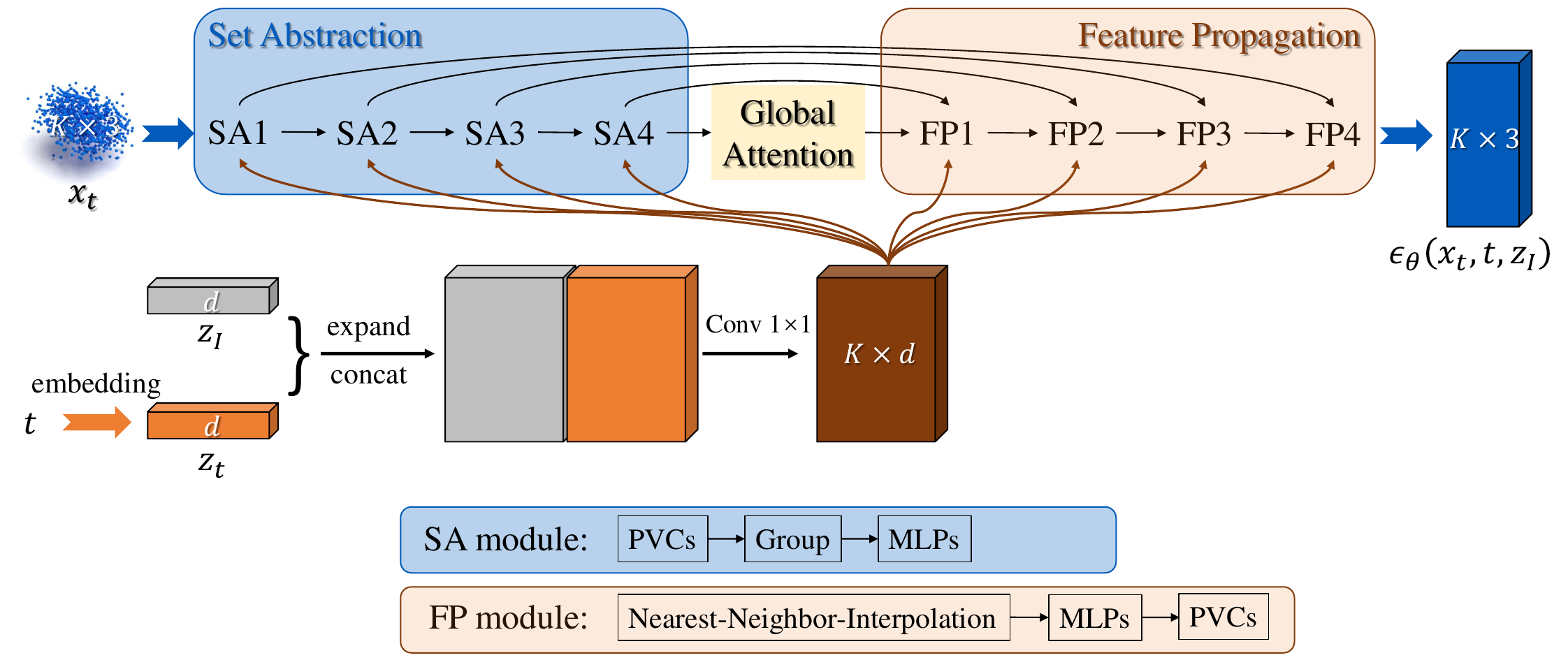}
    \caption{The architecture of the image conditional denoising network $\theta$.}
    \vspace{-4mm}
    \label{fig:supp_theta}
\end{figure*}

\subsection{Image Auto-encoder Pre-training}\label{AE}

A common and straightforward way to condition diffusion models is to compress the cue images into a latent space. Rather than using an encoder pre-trained on public image databases (e.g., ImageNet \cite{deng2009imagenet}) directly map an input image to a latent feature vector that serves as ambiguous conditions from buildings, we instead fine-tune the encoder and train an additional decoder with the building images of the training set. As shown in Fig. \ref{fig:supp_ae}, the overall network can be regarded as an image auto-encoder, which learns to reconstruct the input building images and extracts the features of buildings acting as representative conditions.

Taking an RGB image $I$ of size $\textit{H}\times\textit{W}$ as input, our image auto-encoder employs a ResNet-34 \cite{he2016deep} based encoder to output feature maps of size $\textit{H/32}\times\textit{W/32}$, which is then fed into stacked dilated convolution layers \cite{yu2015multi} (denoted as $D$*). There are four dilated convolution layers with dilation rates of 1, 2, 4, 8, which are stacked in cascade mode and parallel mode to aggregate features from different scales. For the feature maps derived from $D$*, there are two processing ways in the decoding stage. Through transposed convolution layers, the feature maps can be upsampled to $\hat{I}$ that exhibits the same size of the input image $I$. On the other hand, they are linearly projected into a 1-D image embedding $z_I$ with a dimensionality of $d$ via a 1$\times$1 convolution layer and a linear layer.

The image auto-encoder is trained by minimizing $\mathcal{L}_{AE}$ defined as,
\begin{equation}
\mathcal{L}_{AE}=\mathcal{L}_{rec}(I,\hat{I})+\mathcal{L}_{con}(z_I,z_I^a)
\end{equation}
where $\mathcal{L}_{rec}$ is a reconstruction loss between $I$ and $\hat{I}$, $\mathcal{L}_{con}$ is a consistency loss which encourages the embedding $z_I$ of an image $I$ to be as close as possible to the embedding $z_I^a$ of the augmented version $I^a$ of the image $I$. After training, we use the frozen pre-trained image auto-encoder to map the image to an embeddings $z_I$, which acts as the image-dependent conditions for the following diffusion models.

\subsection{Image-conditional Point Cloud Diffusion}\label{Diff}

A conditional diffusion model contains a forward diffusion process (see orange arrows in Fig. \ref{fig:pipeline}) and a denoising diffusion process (see blue arrows in Fig. \ref{fig:pipeline}). Given 3D point cloud $x_0\sim q(x_0)$ of a building with $K$ points, the forward diffusion gradually adds noise by $q(x_t|x_{t-1})$ using a sequence of increasing noise schedules $\beta_t\in\{\beta_1,...\beta_T\}$ where $T$ is the final time step. Assuming $\alpha_t:=1-\beta_t$ and $\bar{\alpha}_t:=\prod_{s=1}^{t}\alpha_s$, noisy point cloud $x_t$ can be represented by,
\begin{equation}
    x_t=\sqrt{\bar{\alpha}_t}x_0+\sqrt{1-\bar{\alpha}_t}\epsilon
\end{equation}
where the time step $t$ is sampled from discrete values $\{1,...,T-1,T\}$. 

The denoising diffusion starts from a random noise tensor $x_T\in\mathbb{R}^{K\times3}$ sampled from a Gaussian prior distribution $p(x_T)\sim\mathcal{N}(0,\mathcal{I})$. The noise is progressively removed by $q(x_{t-1}|x_t,z_I)$, which can be approximated by $p_\theta(x_{t-1}|x_t,z_I)$ parameterized with a denoising network $\theta$. For each denoising step, $\theta$ takes as input noisy point cloud $x_t\in\mathbb{R}^{K\times3}$, time step $t$ and image embedding $z_I\in\mathbb{R}^{d}$. It outputs a noise $\epsilon_\theta(x_t,t,z_I)\in\mathbb{R}^{K\times3}$ and the target is a standard Gaussian noise $\epsilon\sim\mathcal{N}(0,\mathcal{I})$. The denoising loss $\mathcal{L}_{eps}$, which is commonly used as a simplified training objective in DDPMs, is denoted as,
\begin{equation}
\mathcal{L}_{eps}=\left \|\epsilon-\epsilon_\theta(x_t,t,z_I)\right \|^2
\end{equation}

As shown in Fig. \ref{fig:supp_theta}, $\theta$ is basically built on PVCNNs \cite{liu2019point} with two major components, i.e., set abstraction (SA) modules and feature propagation (FP) modules. SA modules typically consist of point-voxel convolutions (PVConvs) and multi-layer perceptrons (MLPs); while FP modules typically consist of nearest neighbor interpolation, MLPs and PVConvs. By using both point-based and voxel-based branches, PVConvs can capture global and local structures of point clouds. Before sending input into the SA or FP modules, the image embedding $z_I$ is concatenated with the temporal embedding (denoted as $z_t$) derived from $t$. Similar to \cite{zhou20213d}, we employ sinusoidal positional embedding \cite{vaswani2017attention} to produce $z_t$ that consists of pairs of sines and cosines with varying frequencies, followed by two linear layers with LeakyReLU activation function. Here, $z_I$ and $z_t$ have the same dimension $d$, and they are concatenated after being expanded to the size of $K\times d$ where $K$ is the number of points. We feed the fused feature map ($K\times2d$) through two convolution layers with LeakyReLU activation function, resulting in a feature map of size $K\times d$ which are then concatenated with point features in SA and FP modules. 

During training, we additionally introduce a weighted building footprint-based regularization strategy. Based on the predicted $\epsilon_\theta$, the desired $x_0$ can be reconstructed by,
\begin{equation}
    \hat{x}_0=\tfrac{1}{\sqrt{\bar{\alpha}_t}}(x_t-\sqrt{1-\bar{\alpha}_t}\epsilon_\theta(x_t,t,z_I))
\end{equation}
Considering the nature of buildings standing on the ground, the reconstructed $\hat{x}_0$ and the target $x_0$ are projected (denoted as $proj$) to the ground, i.e., $z=0$, to obtain footprints $proj(\hat{x}_0)$ and $proj(x_0)$. Then, we adopt a point-based metric $\Omega$ to measure the similarity between two footprints. The regularization loss $\mathcal{L}_{reg}$ can be formulated as,
\begin{equation}
\mathcal{L}_{reg}=\lambda(t)\Omega(proj(x_0),proj(\hat{x}_0))
\end{equation}
where $\lambda$ is the weight depending on time step $t$. When $t$ is close to $T$, $x_t$ could be noisy as $x_T$ is standard Gaussian noise, so $\Omega$ is assigned with lower weight $\lambda$. On the contrary, $\Omega$ is assigned with higher weight $\lambda$ when $t$ is close to 1. 
Overall, the denoising network $\theta$ is optimized by minimizing,
\begin{equation}
\mathcal{L}_\theta=\mathcal{L}_{eps}+\rho\mathcal{L}_{reg}
\end{equation}
where regularization weight $\rho$ balances these two terms.

The overall training process is described in the Algorithm~\ref{alg:train}. Similar to \cite{nichol2022point}, we employ a classifier-free guidance strategy \cite{ho2022classifier} which jointly learns a conditional and an unconditional model. The conditioning image embedding $z_I$ is randomly dropped, thus the conditional output $\epsilon_\theta(x_t,t,z_I)$ is randomly replaced by the unconditional one $\epsilon_\theta(x_t,t,\varnothing)$. 

\begin{algorithm}
\caption{Training of conditional diffusion model}
\label{alg:train}
\begin{algorithmic}
\While{not converge}
\State sample $x_0\sim q(x_0)$, $\epsilon\sim\mathcal{N}(0,\mathcal{I})$
\State sample $t\sim\mathcal{U}(\{1,...,T-1,T\})$
\State $x_t=\sqrt{\bar{\alpha}_t}x_0+\sqrt{1-\bar{\alpha}_t}\epsilon$
\State $\mathcal{L}_{eps}=\left \|\epsilon-\epsilon_\theta(x_t,t,z_I)\right \|^2$
\State $\hat{x}_0=\tfrac{1}{\sqrt{\bar{\alpha}_t}}(x_t-\sqrt{1-\bar{\alpha}_t}\epsilon_\theta(x_t,t,z_I))$
\State $\mathcal{L}_{reg}=\lambda(t)\Omega(proj(x_0),proj(\hat{x}_0))$
\State $\mathcal{L}_\theta=\mathcal{L}_{eps}+\rho\mathcal{L}_{reg}$
\State update model parameter $\theta$ with $\nabla_\theta\mathcal{L}_\theta$
\EndWhile
\end{algorithmic}
\end{algorithm}

During sampling, the denoising diffusion begins from a Gaussian noise (i.e., $x_T\sim\mathcal{N}(0,\mathcal{I})$), and denoises with the output of network $\theta$ step by step. $x_{T-1}$ can be predicted by,
\begin{equation}
    x_{t-1}=\tfrac{1}{\sqrt{\alpha_t}}(x_t-\tfrac{1-\alpha_t}{\sqrt{1-\bar{\alpha}_t}}\epsilon_{guided}(x_t,t,z_I))+\sigma_t \textbf{z}
\end{equation}
where $t$ begins from $T$ to 1, and \textbf{z} is sampled from standard Gaussian distribution when $t>1$.  Leveraging a guidance scale $\gamma$, the guided noise output is,
\begin{equation}
 \epsilon_{guided}:=(1+\gamma)\epsilon_\theta(x_t,t,z_I)-\gamma\epsilon_\theta(x_t,t,\varnothing)   
\end{equation}
Eventually, the desired $x_0$ can be sampled when $t=1$.

\subsection{Point Cloud Upsampler Diffusion}

For point cloud base diffusion, the key objective is generating low-resolution point clouds ($K$ points) that could coarsely capture the overall structure of buildings. We train another diffusion model conditioned on the image embedding $z_I$ derived by a frozen pre-trained auto-encoder and the low-resolution point cloud inferred by base diffusion. The goal lies in generating high-resolution point cloud with fine-grained structure. Thus, the second diffusion model is called upsampler diffusion.

Our upsampler diffusion leverages similar architecture as our base diffusion model. Assuming the desired point cloud $x_0\in\mathbb{R}^{N\times3}$ consists of $N$ points ($N>K$), we randomly sample a noise tensor $x_T\in\mathbb{R}^{N\times3}$ from a Gaussian prior distribution $p(x_T)\sim\mathcal{N}(0,\mathcal{I})$. During training, the denoising network $\theta$ takes as input $K$ points (i.e., low-resolution point cloud) and $(N-K)$ points sampled from noisy $x_t$, time step $t$ and conditioning image embedding $z_I$. At each step, the first $K$ of $N$ points sampled by  $\theta$ is replaced by the low-resolution point cloud, and the updated $N$ points are used as input in the next time step. In short, to arrive at $N$ points, our upsampler conditions on $K$ points and denoises the rest $(N-K)$ points.

\section{Experiments}

\subsection{Datasets}

To validate the performance of the proposed method, we create two datasets, \textbf{BuildingNet-SVI} and \textbf{BuildingNL3D}, providing thousands of image-3D pairs of buildings. 

\vspace{1mm}
\noindent
\textbf{BuildingNet-SVI.} Built on BuildingNet \cite{selvaraju2021buildingnet} dataset that covers a variety of synthetic 3D building models (e.g., churches, houses, and office buildings), we collect the corresponding single-view RGB synthetic images of buildings and obtained 406 image-3D pairs after quality checks w.r.t. completeness and consistency. Besides, we manually annotate the foreground object (i.e., individual building) at the pixel level, and crop each image centered around the building. Each building point cloud has 100,000 uniformly distributed 3D points. We follow the official splitting rules from BuildingNet \cite{selvaraju2021buildingnet}, thus 321 and 85 image-3D pairs are used for training and testing, respectively. 

\vspace{1mm}
\noindent
\textbf{BuildingNL3D.} We collect 2,769 pairs of aerial RGB images and Airborne Laser Scanning (ALS) point clouds of buildings which are located in the urban area of a city in the Netherlands. Unlike synthetic images that have individual buildings and relatively clean backgrounds, aerial images usually face more challenges such as multiple buildings appearing in a single image. Hence, the buildings are manually labeled so that only one building of interest appears in each image. Raw ALS point clouds are normalized from their real geographic coordinates to $xyz$ coordinates within the range $[-1, 1]^3$. The dataset is divided into 2,171 image-3D training pairs and 598 test pairs according to tile-based splitting rules where buildings in the training and test sets are not repeated. 

\subsection{Evaluation Metrics}

We adopt Chamfer distance (CD) \cite{fan2017point}, Earth mover's distance (EMD) \cite{rubner2000earth} and F1-Score \cite{knapitsch2017tanks} to evaluate the pair-wise similarity between a generated building and its reference building. Specifically, CD and EMD are multiplied by $10^2$, and the threshold $\tau$ of F1 is set as 0.001. The point clouds are normalized into $[-1, 1]^3$ before calculating with these metrics. The visualization of 3D point clouds is achieved by using Mitsuba Renderer \cite{render}. 

\begin{figure*}[t!]
    \centering
    \includegraphics[width=0.88\linewidth]{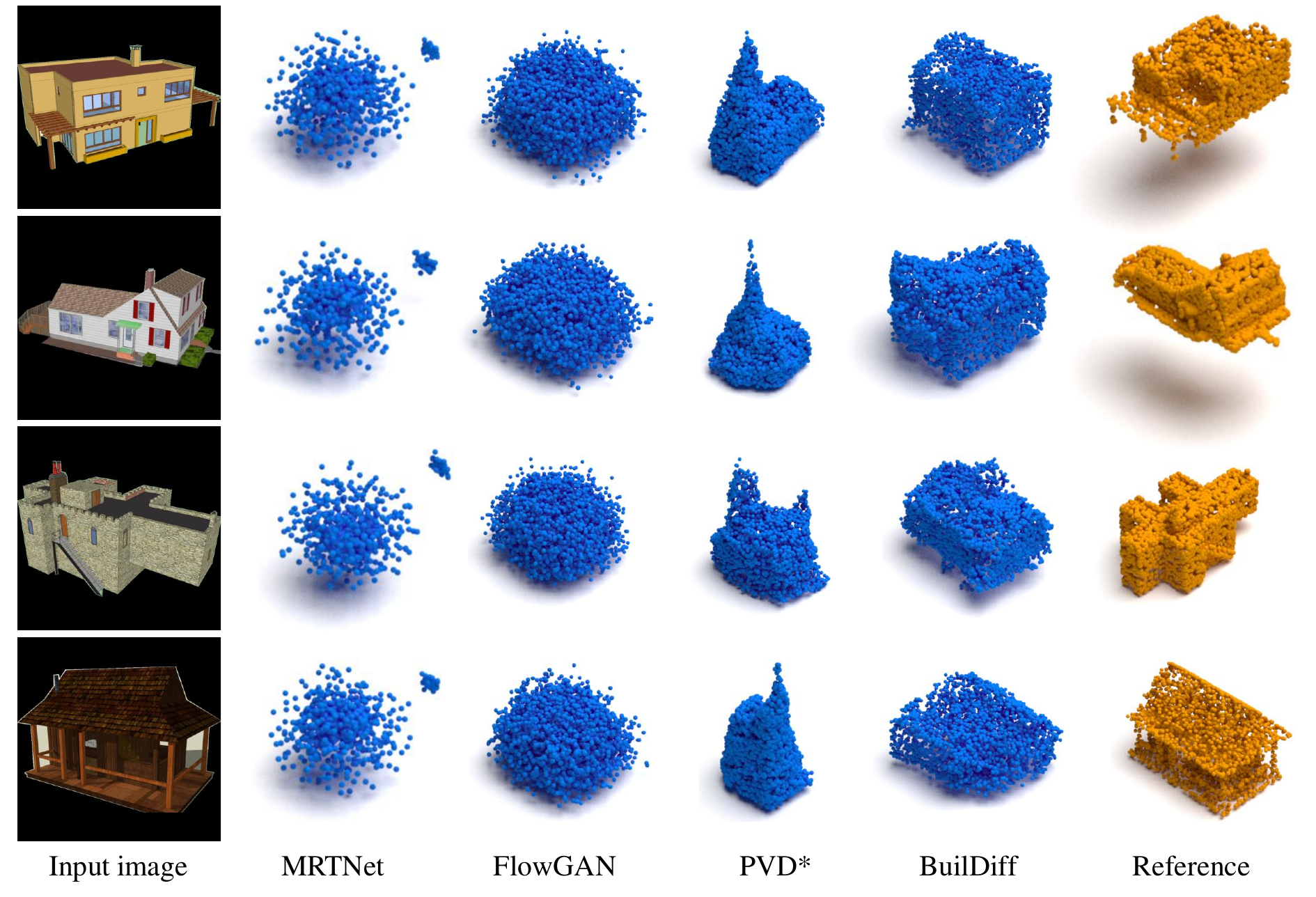}
    \caption{
    Qualitative comparison of different 3D point cloud generation models on the BuildingNet-SVI dataset. The reference and predicted point clouds are colored by orange and blue, respectively.
    }
    \vspace{-3mm}
    \label{fig:result_bui}
\end{figure*}

\begin{figure*}[ht]
    \centering
    \includegraphics[width=0.88\linewidth]{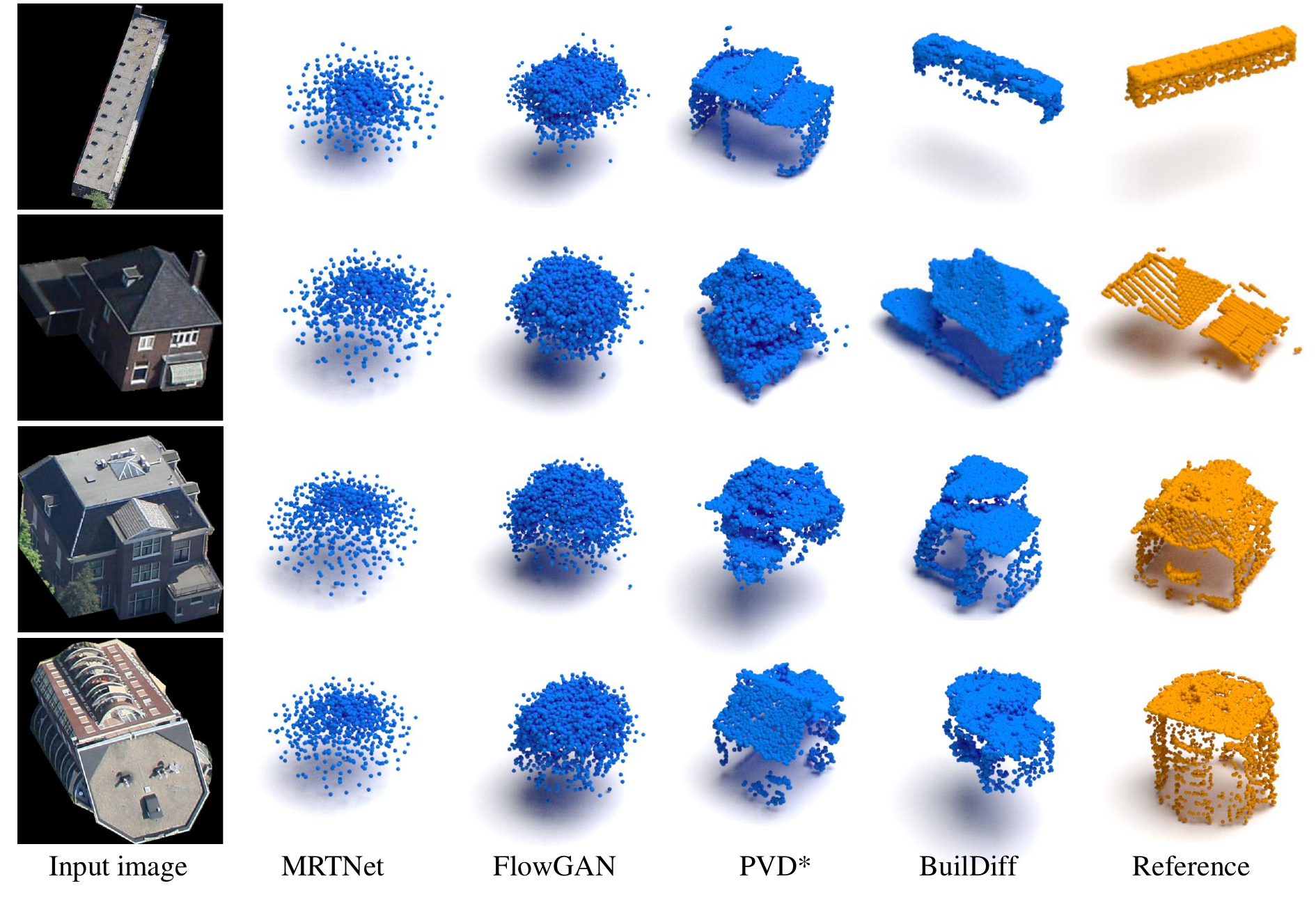}
    \caption{Qualitative comparison of different 3D point cloud generation models on the BuildingNL3D dataset. The reference and predicted point clouds are colored by orange and blue, respectively.}
    \vspace{-3mm}
    \label{fig:result_ens}
\end{figure*}

\subsection{Implementation Details}

The models are implemented with PyTorch \cite{paszke2019pytorch} on a NVIDIA A40 GPU with 45GB  memory. The batch size is set as 8. The images are resized into 1024$\times$1024 pixels, i.e., $H=W=$ 1024. Each 3D point cloud has normalized 100,000 points for representing the shape of the building.

\vspace{1mm}
\noindent
\textbf{Image auto-encoder details.}
We adopt augmentation techniques including image rotation with an angle of 90\degree and color shifting in the Hue-Saturation-Value (HSV) space ranging from -255 to 255. These are based on the consideration that buildings may appear upside down in the images and building images could exhibit large color variations. The dimensionality $d$ of image embedding $z_I$ is 128. We train the auto-encoder for 30 epochs and use the Adam optimizer with a learning rate 0.0002. The image reconstruction loss $\mathcal{L}_{rec}$ and the embedding consistency loss $\mathcal{L}_{con}$ are achieved by mean squared error (MSE). During training the following image-conditional diffusion models, we freeze the pre-trained autoencoder to derive $z_I$.

\vspace{1mm}
\noindent
\textbf{Base diffusion details}.
Given 100,000 points per building, we randomly sample $K=$ 1024 points (representing low-resolution point clouds) for training and testing. We set $\beta_0=$ 0.0001, $\beta_T=$ 0.02 and linearly interpolate other $\beta$s. Similar to existing works \cite{zhou20213d,zeng2022lion}, the total number of time steps is set as $T=$ 1000 for the base diffusion. The dimensionality of temporal embedding is $d=$ 128. Chamfer distance \cite{fan2017point} is employed as the point-based distance metrics $\Omega$ to measure the similarity between two projections. The regularization weight $\rho$ is set as 0.001. Regarding the classifier-free guidance strategy, we employ drop probability 0.1 for the training phase, and guidance scale $\gamma=$ 4 for the sampling phase. The diffusion model is trained for 700 epochs and optimized by Adam with a learning rate 0.0002. Specifically, $\lambda(t)$ is defined as,

$\lambda(t) = \begin{cases}
1, & t=1 \\
0.75, & 1<t\leq\frac{1}{4}T \\
0.50, & \frac{1}{4}T<t\leq\frac{1}{2}T \\
0.25, & \frac{1}{2}T<t\leq\frac{3}{4}T \\
0, & \frac{3}{4}T<t\leq T \\
\end{cases}$

\vspace{1mm}
\noindent
\textbf{Upsampler diffusion details}.
For the training and testing phases, we randomly sample $N=$ 4096 points per building. The denoising network of the upsampler is trained for 200 epochs and the total number of time steps $T$ is set as 500. 

\subsection{Quantitative and Qualitative Results}

To demonstrate the performance of the proposed method, BuilDiff is compared with several 3D point cloud generation approaches based on deep generative models. The comparison methods are MRTNet \cite{gadelha2018multiresolution} which leverages VAEs for single image 3D reconstruction, FlowGAN \cite{wei2022flow} which reconstructs 3D point cloud from a single image by combining the advantages of NFs and GANs, and PVD \cite{zhou20213d} which use diffusion models for 3D point cloud generation. As PVD is originally designed for unconditional point cloud synthesis, we extend its framework by introducing a global embedding extracted from images as an auxiliary input of denoising diffusion models, thus we note it as PVD*. Our method differs from PVD* in three main aspects: (1) an auto-encoder is pre-trained to extract more representative conditions $z_I$ to avoid ambiguous guidance; (2) a weighted footprint-based regularization strategy is introduced to the training of denoising network $\theta$; (3) a point cloud upsampler is used to produce high-resolution point clouds of buildings.

The implementation of the comparison methods is based on their officially released code. All the models are trained and evaluated on the BuildingNet-SVI dataset and BuildingNL3D dataset, the quantitative results of which are shown in Table \ref{table:result@BUI} and Table \ref{table:result@ENS}. Here, we sample 4,096 points per shape to ensure a fair comparison. 

Regarding the synthetic data, it can be seen from Table \ref{table:result@BUI} that BuilDiff achieves the best EMD and F1 as well as the second-best CD. Unlike common 3D objects (e.g., ShapeNet) that other methods focus on, buildings have more complex structures that are not symmetrical. Such natures make it difficult to learn the shape of buildings in canonical object space. Correspondingly, directly using VAEs or GANs to constrain the network output may prevent the network from learning valuable information. 

In Table \ref{table:result@ENS}, BuilDiff reports the lowest EMD, second-highest F1, and comparable CD with the baseline method PVD*. Regarding other methods, MRTNet directly employs CD loss for training, it obtains low CD while performs poorly on other metrics. FlowGAN obtains the highest F1 but inferior EMD, demonstrating its preference for generating uniform global distributed points while ignoring local details. As EMD is stricter with local quality \cite{wu2021density}, our results show robust performance in terms of both global shapes and local structures.

\begin{table}
\begin{center}
\begin{tabular}{c|c c c}
\hline
Methods & CD$\downarrow$ & EMD$\downarrow$ & F1$\uparrow$\\
\hline
MRTNet \cite{gadelha2018multiresolution} & 6.11 & 49.07 & 6.89 \\
FlowGAN \cite{wei2022flow} & \textbf{2.00} & 21.21 & \underline{21.17}\\
PVD \cite{zhou20213d} * & 6.18 & \underline{16.08} & 20.02\\
BuilDiff & \underline{3.14} & \textbf{10.84} & \textbf{21.41} \\
\hline
\end{tabular}
\end{center}
\caption{Quantitative results on BuildingNet-SVI dataset. The best results are indicated in \textbf{bold} and the second best are marked with \underline{underlines}.}
\vspace{-7mm}
\label{table:result@BUI}
\end{table}

\begin{table}
\begin{center}
\begin{tabular}{c|c c c}
\hline
Methods & CD$\downarrow$ & EMD$\downarrow$ & F1$\uparrow$\\
\hline
MRTNet \cite{gadelha2018multiresolution} & \underline{2.84} & 44.06 & 5.18 \\
FlowGAN \cite{wei2022flow} & \textbf{2.33} & 24.06 & \textbf{22.26}\\
PVD \cite{zhou20213d} *& 5.69 & \underline{14.74} & 13.01\\
BuilDiff & 3.81 & \textbf{10.43} & \underline{22.08} \\
\hline
\end{tabular}
\end{center}
\caption{Quantitative results on BuildingNL3D dataset. The best results are indicated in \textbf{bold} and the second best are marked with \underline{underlines}.}
\vspace{-5mm}
\label{table:result@ENS}
\end{table}

\begin{table}[http]
\begin{center}
\begin{adjustbox}{width=0.49\textwidth}
\begin{tabular}{c|c|c|c|c c c}
\hline
pts & pre-train AE & regularization & upsampler & CD$\downarrow$ & EMD$\downarrow$ & F1$\uparrow$ \\
\hline
1024 & $\times$ & $\times$ & - & 4.523 & 12.244 & 14.042\\
1024 & \checkmark & $\times$ & - & 4.759 & 12.847 & \textbf{14.631}\\
1024 & \checkmark & \checkmark & - & \textbf{3.556} & 11.747 & 13.889\\
1024 & \checkmark & \checkmark\dag & - & 3.827 & \textbf{10.682} & 13.376\\
\hline
4096 & $\times$ & $\times$ & - & 6.181 & 16.075 & 20.022\\
4096 & \checkmark & \checkmark\dag & - & 3.678 & 15.127 & 21.128\\
4096 & \checkmark & \checkmark\dag & \checkmark & \textbf{3.142} & \textbf{10.840} & \textbf{21.413}\\
\hline
\end{tabular}
\end{adjustbox}
\end{center}
\caption{Ablation studies on the BuildingNet-SVI dataset. The best results are shown in \textbf{bold}.}
\vspace{-3mm}
\label{table:ablation@BUI}
\end{table}

The qualitative results of several generated 3D buildings are shown in Fig.~\ref{fig:result_bui} and Fig.~\ref{fig:result_ens}. MRTNet and FlowGAN yield unsatisfactory results, which demonstrates their poor generalization ability in 3D building reconstruction. In contrast, diffusion-based methods perform better in capturing the shape of buildings. Due to the simple conditioning strategy used by PVD*, the generated 3D buildings are inconsistent with the input condition images. By the proposed conditioning schemes, BuilDiff decreases the diversity of diffusion models and increases the quality of generated buildings. Furthermore, it can be observed from the last row of Fig.~\ref{fig:result_ens} that our method demonstrates the robustness even when buildings appear upside down in the input image.

\subsection{Ablation Study}

We verify the effectiveness of the proposed components, i.e., pre-training image auto-encoder,  weighted footprint-based regularization, and point cloud upsampler. The comparison results are shown in Table \ref{table:ablation@BUI} and Table \ref{table:ablation@ENS}. The first four rows are tested on 1,024 points in which our upsampler is not used. We sequentially compare the effects of pre-training image auto-encoder, footprint-based regularization, and weighted footprint-based regularization that is denoted with \dag. The last three rows are tested on 4,096 points. 

\begin{table}
\begin{center}
\begin{adjustbox}{width=0.49\textwidth}
\begin{tabular}{c|c|c|c|c c c}
\hline
pts & pre-train AE & regularization & upsampler & CD$\downarrow$ & EMD$\downarrow$ & F1$\uparrow$ \\
\hline
1024 & $\times$ & $\times$ & - & 11.109 & 24.027 & 5.180\\
1024 & \checkmark & $\times$ & - & 6.406 & 16.770 & 8.531\\
1024 & \checkmark & \checkmark & - & \textbf{5.680} & 15.813 & 8.928\\
1024 & \checkmark & \checkmark\dag & - & 5.766 & \textbf{15.625} & \textbf{9.601}\\
\hline
4096 & $\times$ & $\times$ & - & 5.691 & 14.735 & 13.012\\
4096 & \checkmark & \checkmark\dag & - & 4.046 & 13.373 & 15.227\\
4096 & \checkmark & \checkmark\dag & \checkmark & \textbf{3.810} & \textbf{10.427} & \textbf{22.081}\\
\hline
\end{tabular}
\end{adjustbox}
\end{center}
\caption{Ablation studies on the BuildingNL3D dataset. The best results are shown in \textbf{bold}.}
\vspace{-5mm}
\label{table:ablation@ENS}
\end{table}

\begin{figure}[t!]
    \centering
    \includegraphics[width=\linewidth]{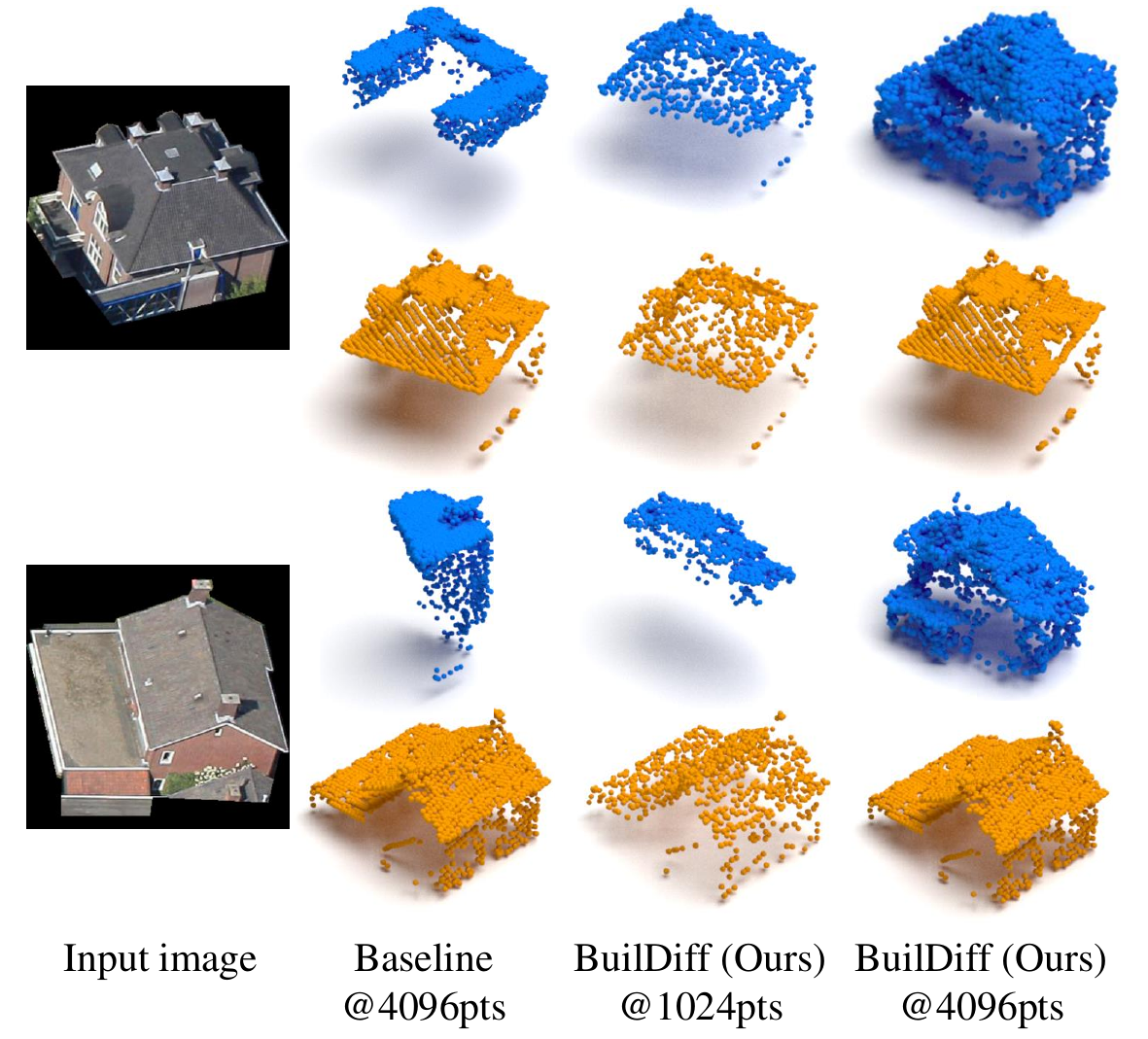}
    \caption{The illustration of the effect of our upsampler diffusion. The reference and predicted point clouds are colored by orange and blue, respectively.}
    \vspace{-4mm}
    \label{fig:supp_illu}
\end{figure}

First, it can be observed that our image auto-encoder greatly increases the performance on real data from BuildingNL3D, which means that it could alleviate the domain gap between real-world building images and common images learned by image classification networks. Such expressive condition benefits from multi-scale features and latent consistency performed by image augmentation. Second, the model is further improved by introducing building footprint-based regularization in terms of CD and EMD. The weighted regularization guides the network focus on later denoising steps. Finally, it can be seen from the last two rows that our BuilDiff performs better than the baseline (i.e., pre-train AE + regularization, without upsampling), demonstrating the effect of our upsampler diffusion. Visualizations can be found in Fig. \ref{fig:supp_illu}. Baseline method sometimes has difficulty focusing on the basic structure of the buildings due to too many points to consider, leading to 3D buildings that are completely unrelated to the input images. In contrast, BuilDiff employs a base diffusion model that coarsely identifies the overall structure of buildings, and then uses an upsampler diffusion to derive high-resolution point clouds conditioned on low-resolution point clouds and input images. Our method can even generate more complete buildings than the reference point clouds, for example, the final results in the last column contain realistic building facades.

\section{Conclusions}

In this paper, we present a diffusion-based method \textbf{BuilDiff} for generating 3D point clouds of buildings from single general-view images. To control the diffusion models generating 3D shape consistent with the input image, an image embedding is derived by pre-training a CNN-based image auto-encoder, which extracts multi-scale features of buildings and constrains the latent consistency using augmentation. Then, a conditional denoising diffusion network which takes as input the image embedding and learns to gradually remove the noise from Gaussian noise distribution assisted by weighted building footprint-based regularization. A point cloud upsampler diffusion is finally leveraged to produce high-resolution point clouds conditioned on the low-resolution point clouds sampled from the base diffusion. The effectiveness of the proposed method has been demonstrated by the experimental results on both synthetic and real-world scenarios. We believe our work could bridge the rapidly developing generative modeling techniques and the urgent problem of 3D building generation.

{\small
\bibliographystyle{ieee_fullname}
\bibliography{egbib}
}

\end{document}